\documentclass[letterpaper]{article} % DO NOT CHANGE THIS
\usepackage[preprint]{aaai2027}  % DO NOT CHANGE THIS
\usepackage[hyphens]{url}  % DO NOT CHANGE THIS
\usepackage{graphicx} % DO NOT CHANGE THIS
\usepackage{natbib}  % DO NOT CHANGE THIS AND DO NOT ADD ANY OPTIONS TO IT
\usepackage{caption} % DO NOT CHANGE THIS AND DO NOT ADD ANY OPTIONS TO IT
\usepackage{algorithm}
\usepackage{algorithmic}

\usepackage{newfloat}
\usepackage{listings}
\DeclareCaptionStyle{ruled}{labelfont=normalfont,labelsep=colon,strut=off} % DO NOT CHANGE THIS
\floatstyle{ruled}
\newfloat{listing}{tb}{lst}{}
\floatname{listing}{Listing}

\usepackage{booktabs}
\usepackage{amsmath}
\usepackage{amssymb}
\usepackage{subcaption}

\usepackage{tabularx}
\newcolumntype{C}{>{\centering\arraybackslash}X}  % 居中的可伸缩列

\title{Exposing the Long-tail in Embodied Urban Navigation via Scalable Learning from In-the-Wild Videos}
\author {
    Bingyi Xia\textsuperscript{\rm 2},
    Han Bao\textsuperscript{\rm 2},
    Zhewei Chen\textsuperscript{\rm 2}, 
    Hanjing Ye\textsuperscript{\rm 2}, 
    Jingwen Yu\textsuperscript{\rm 3}, 
    Yuhan Pang\textsuperscript{\rm 2}, \\
    Wenjun Xu\textsuperscript{\rm 1}\corresponding,  
    Jiankun Wang\textsuperscript{\rm 2}\corresponding
}
\affiliations{
    \textsuperscript{\rm 1}Peng Cheng Laboratory\\
    \textsuperscript{\rm 2}Southern University of Science and Technology\\
    \textsuperscript{\rm 3}The Hong Kong University of Science and Technology\\
     xiaby2020@mail.sustech.edu.cn, xuwenjunwendy@gmail.com, wangjk@sustech.edu.cn
}

\begin{document}

\maketitle

\begin{abstract}
Learning embodied urban navigation policies from real-world data is constrained by the cost of task-specific data collection and the limited coverage of rare yet safety-critical scenarios. To address these challenges, we present a scalable framework for learning point-goal urban navigation from web-scale in-the-wild egocentric videos while systematically exposing its long tail. The framework automatically annotates uncurated web videos with metric trajectories and structured navigation semantics, which are then used to train a vision-language-action policy for interpretable navigation planning. We characterize the long tail based on model performance and the distribution of perception–motion patterns, and employ reflection-based analysis to diagnose recurring failure modes. Experiments on web-video data and real-world urban navigation tasks demonstrate effective knowledge transfer from unconstrained videos and reveal coherent long-tail structures beyond aggregate navigation performance.
\end{abstract}


\section{Introduction}
\label{sec:introduction}

\begin{figure*}[!t]
\centering
    \includegraphics[width=1.00\linewidth]{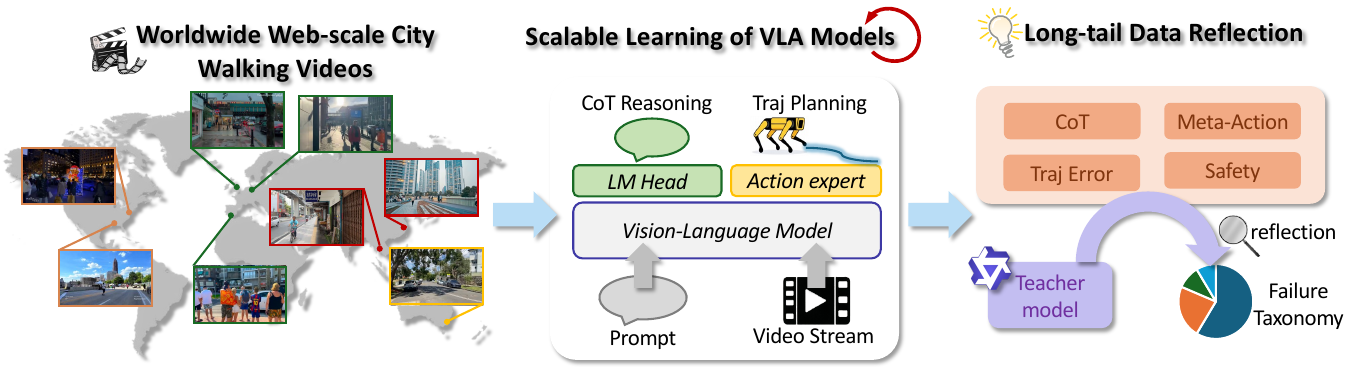}
    \caption{
Overview of the proposed framework for scalable urban navigation VLA models. Our approach first collect worldwide web-scale street-walking videos as the dataset. A VLA model is then trained to perform end-to-end urban navigatioin planning. Finally, the reflection driven by a teacher model is introduced to analyze failure taxonomies.
    }
    \label{fig:cover}
\end{figure*}

Urban navigation is a fundamental capability for embodied agents such as delivery robots operating in public spaces. 
Unlike structured indoor environments or lane-governed roads, open urban spaces involve dense pedestrian interactions, irregular traversable regions, and implicit social conventions. 
Traditional modular pipelines rely on predefined representations and rules, whereas recent navigation foundation models adopt end-to-end frameworks that generalize better to such scenarios \cite{omnivla,socialnav}.

Following the scaling paradigm, foundation models for embodied urban navigation have grown in size, increasing the need for large-scale and diverse supervision.
Their training process commonly relies on simulated data or collected human demonstrations \cite{s2e, socialnav}. 
However, publicly available teleoperated datasets remain limited in scale, while simulated data can be generated efficiently but remains subject to the sim-to-real gap. 
More fundamentally, they are typically constructed within predefined tasks and carefully designed scenarios, inevitably limiting the coverage of open-world situations. 
Therefore, they are poorly suited to systematically exposing unforeseen tail patterns or revealing where models fail across a broader empirical distribution. 
Further progress, however, depends not only on adding more samples but also on whether task-specific data cover the situations that constrain model performance.

Long-tail coverage therefore becomes a central concern: which navigation experiences remain missing or underrepresented and have become the bottleneck to further improvement?
In-the-wild egocentric videos offer an alternative by capturing human navigation across naturally occurring environments at scale. 
Prior work has demonstrated that such videos can provide effective supervision for navigation policy learning \cite{roomtour3d, citywalker, urbannav}. 
More importantly, their broad empirical distribution supports the joint analysis of rare perception--motion patterns and model-dependent hard cases, making in-the-wild videos not only a scalable source of training data but also an empirical basis for identifying the task-data bottlenecks and failure modes that limit further scaling.

In this work, we present a scalable framework for learning point-goal urban navigation from in-the-wild egocentric videos and systematically exposing its long tail. 
We curate about 500 hours of street-walking videos spanning 30 cities and develop an automated pipeline to recover metric trajectories and generate structured navigation chain-of-thought annotations, yielding more than 500K samples. 
These data are used to train \textbf{WILD-Nav}, our proposed reasoning-aware vision-language-action model for interpretable point-goal urban navigation planning.
Beyond aggregate performance, we characterize the long tail by jointly analyzing the distributional rarity of perception--motion patterns and model-dependent difficulty, and employ reflection-based analysis to attribute recurring failures to different reasoning and planning stages.
Experiments on video and real-world navigation datasets demonstrate effective knowledge transfer and reveal coherent long-tail structures without relying on predefined scenario taxonomies.
We further organize the identified rare and hard episodes, together with reflection-based failure attribution, into a long-tail benchmark \textbf{WILD-LongTail} for consistent evaluation.

Our contributions are threefold:
\begin{itemize}
    \item We develop a scalable pipeline that annotates egocentric in-the-wild videos of street-walking with trajectories and structured reasoning supervision for training urban navigation VLA models.
    \item We construct a web-scale dataset of over 500K urban navigation samples and provide baseline models for benchmarking policy learning and evaluation. 
    \item We characterize the long-tail through model-dependent difficulty and distributional rarity in perception--motion patterns, with privileged reflection for failure attribution.
\end{itemize}
\section{Related Work}

\subsection{Scalable Data for Navigation}

Navigation VLA models learn from multiple sources, including robot-collected demonstrations, simulation, and egocentric human videos. ~\cite{s2e, socialnav, urbanvla}. 
These sources provide complementary benefits. 
Real-robot data are limited by expensive collection and narrow geographic coverage~\cite{sit, gnd}. 
Although simulation enables closed-loop policy training with accurate labels, it retains a sim-to-real gap and limited predefined rules. 
SimWorld, for instance, is constructed from only 200 navigation episodes~\cite{simworld}. 
Egocentric videos provide a scalable alternative by recording diverse human experiences in naturally occurring environments~\cite{egovla}. 
Prior work has used such videos to learn navigation behaviors and spatial grounding~\cite{lelan, lookout, sawbench}. 

Recent studies have leveraged Internet-sourced data to scale pretraining datasets to thousands of hours~\cite{egoscale, jala}.
More closely related to navigation, FLAME and CityNav exploit online maps and street-view imagery to learn sparse route decisions, while RoomTour3D converts Internet-sourced videos into candidate-view selection tasks as training data for indoor VLN~\cite{flame, citynav, roomtour3d}. 
However, these methods only supply high-level decisions rather than executable trajectories, which cannot be directly employed by urban navigation VLA models. 
Citywalker recovers trajectories from human walking videos shot in NewYork using visual odometry. 
However, substantial scene repetition leaves only approximately 200 hours of effective data~\cite{citywalker}. 
VEGA generates local 3D geometry from individual video frames and trajectories by a model-based planner, focusing primarily on geometric safety and 3D grounding~\cite{vega}.
These studies establish in-the-wild videos as scalable training data, but their natural distributions have received limited attention as a basis for systematically discovering long-tail patterns and model failures.

\subsection{VLA Reasoning for Navigation}

Structured textual annotations have been shown to strengthen the reasoning capabilities for navigation~\cite{vlnvideo, walkgpt}. 
In autonomous driving, AutoDrive-P3 structures reasoning into perception, prediction, and planning, enabling more interpretable action planning~\cite{autodrivep3}. 
Comparable reasoning supervision remains less developed for urban navigation learned from Internet-sourced videos. 
CityWalker only provides reconstructed trajectories, while VEGA provides object-goal labels rather than navigation rationales~\cite{citywalker,vega}. 
Recent urban navigation models introduce additional task-specific language instructions based on Citywalker.
UrbanNav uses landmark-grounded instructions, while SocialNav emphasizes social traversability~\cite{urbannav, socialnav}. 
However, these annotations do not provide a general reasoning structure for urban navigation. 
Furthermore, AutoDrive-R2 demonstrates that self-reflection can validate and correct planned trajectories~\cite{autodriver2}. 
Inspired by them, our work aligns actions with a structured perception--analysis--planning reasoning chain and uses reflection to trace long-tail failures to specific reasoning stages.
As summarized in Table~\ref{tab:dataset}, it provides substantially greater scale and coverage than existing urban navigation datasets, together with structured navigation reasoning annotations.

\subsection{Long-Tail Discovery}
\label{sec:related_work_longtail}
Most long-tail studies identify tail data through the low empirical frequencies of predefined classes or semantic concepts \cite{head2tail,neighbors}. 
Object-category frequency alone, however, does not fully capture challenging urban navigation situations determined by joint perception--motion patterns. 
Moreover, distributional rarity and model-dependent difficulty can overlap but need not coincide, while average performance may conceal high conditional errors on particular subpopulations \cite{sagawa2020groupdro}. 
Rare Example Mining offers a related distinction by defining rareness through low feature-space density within a semantic category and filtering intrinsically difficult examples using task-specific observation-quality rules \cite{rareexample}. 
Existing studies of long-tail in navigation and autonomous driving commonly rely on predefined corner-case taxonomies \cite{impromptu, wode2e, sidewalk}. 
Such taxonomies depend on prior knowledge of which situations are likely to be difficult. 
However, urban navigation lacks comparably mature long-tail taxonomies and broad natural-distribution benchmarks from which they can be derived. 
To address this gap, we use large-scale in-the-wild navigation videos as an empirical testbed and independently identify distributionally rare perception--motion patterns and high-error cases, separating data-coverage gaps from model weaknesses. 
Finally, reflection-based methods suggest that reflective CoT and structured diagnostic reports can expose factors underlying reasoning and planning failures \cite{srpo,elfvla}. 
We accordingly apply privileged reflection to characterize recurring failure attribution among the mined hard cases.

\section{Learning from In-the-Wild Videos}
\label{sec:web_video_learning}

\begin{figure}[t]
\centering
\includegraphics[width=0.4\textwidth]{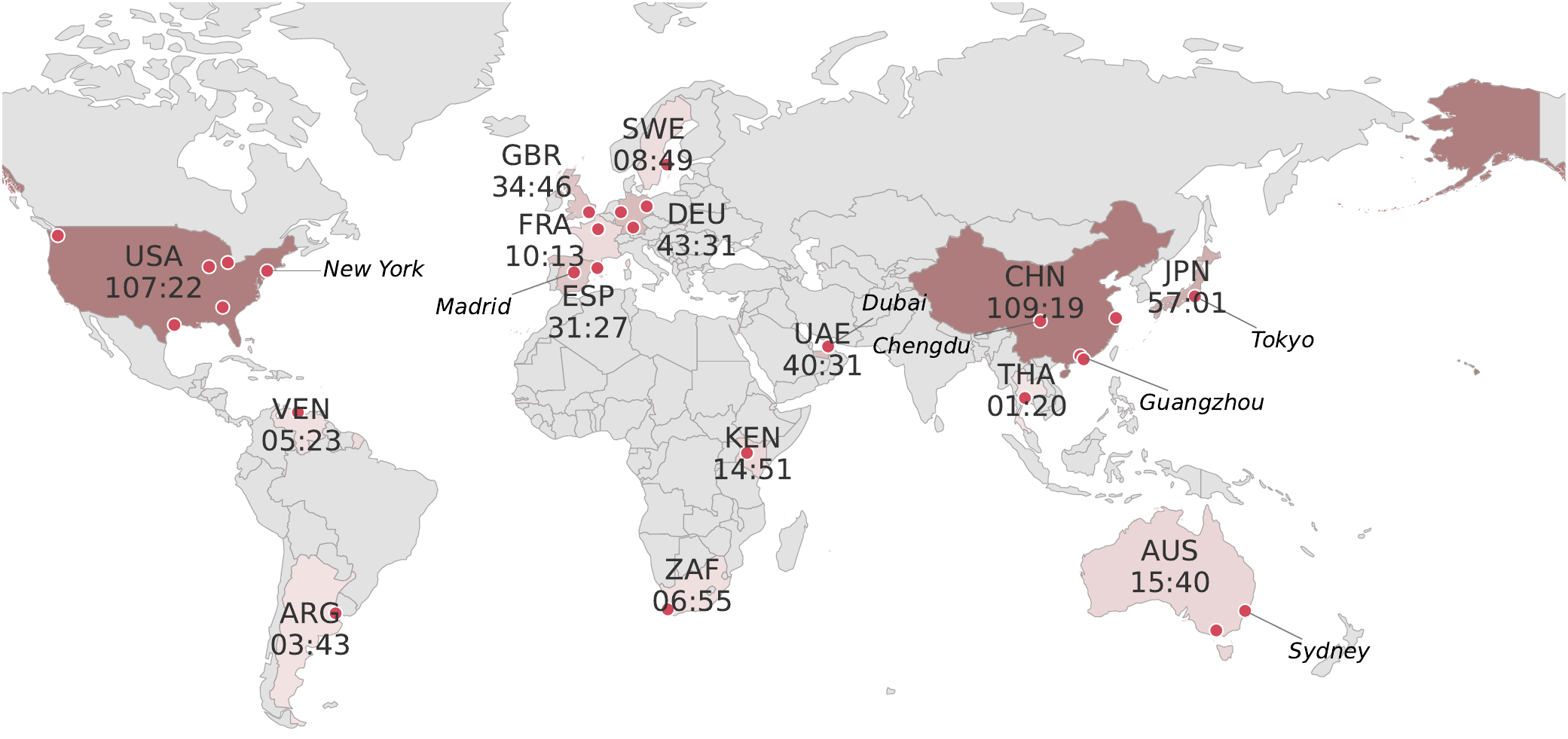} % Reduce the figure size so that it is slightly narrower than the column.
\caption{Geographic distribution of our in-the-wild street-walking video collection.}
\vspace{-0.2cm}
\label{fig:video_coverage}
\end{figure}

\begin{table}[t]
\centering
\caption{Comparison of urban navigation datasets based on real-world collection and in-the-wild videos.}
\label{tab:dataset}
\footnotesize
\setlength{\tabcolsep}{3pt}

\begin{tabular}{l c c c}
\toprule
\textbf{Dataset} & \textbf{Duration} & \textbf{Locations} & \textbf{Annotation} \\
\midrule
SCAND   & 8.7 h & 1  & -  \\
SiT     & 0.4 h & 1  & 3D tracks and semantic maps  \\
GND     & 11 h & 10  & Semantic map   \\
\midrule
Citywalker & 200 h & 1  & Caption and instruction   \\
Ours & 500 h & 30 & Navigation reasoning CoT   \\
\bottomrule
\end{tabular}

\vspace{-0.5cm}
\end{table}

We train WILD-Nav using a geographically broad collection of internet-sourced egocentric videos. 
Our data pipeline converts these videos into urban navigation episodes. 
The pipeline consists of trajectory reconstruction, structured reasoning annotation, and multi-task supervised learning.

\subsection{Problem Formulation}

We consider the foundational problem of local navigation planning in outdoor urban environments. 
Our objective is to learn a real-time, end-to-end policy that plans short-horizon actions without access to a pre-built map or other environmental priors. 
At time step $t$, the agent receives the current monocular RGB observation $I_t \in \mathbb{R}^{H \times W \times 3}$ and an observation history $\mathcal{I}^{\mathrm{hist}}_t = \{I_{t-4},\ldots,I_{t-1}\}$. 
The corresponding short-range localization history is represented by the trajectory vector $\boldsymbol{\tau}^{\mathrm{hist}}_t = [\mathbf{p}_{t-4},\ldots,\mathbf{p}_t]$, where $\mathbf{p}_i\in\mathbb{R}^{2}$. All positions are expressed in the egocentric ground-plane coordinate system. 

Conditioned on a local point goal $\mathbf{g}_t\in\mathbb{R}^{2}$, the policy plans a scale-consistent short-horizon trajectory represented by 5 discrete waypoints, $\hat{\boldsymbol{\tau}}^{\mathrm{plan}}_t=[\hat{\mathbf{p}}_{t+1},\ldots,\hat{\mathbf{p}}_{t+5}]$.
In addition to the executable waypoint plan, WILD-Nav predicts a high-level meta-action $a_t$ and generates a structured navigation rationale $c_t$, following recent reasoning-aware navigation formulations \cite{socialnav, autodriver2}. 
During training, $c_t$ is supervised using structured rationale annotations. 
The complete policy is formulated as
\begin{equation}
    \left(
        c_t,
        a_t,
        \hat{\boldsymbol{\tau}}^{\mathrm{plan}}_t
    \right)
    =
    \pi_{\theta}
    \left(
        I_t,
        \mathcal{I}^{\mathrm{hist}}_t,
        \boldsymbol{\tau}^{\mathrm{hist}}_t,
        \mathbf{g}_t
    \right).
    \label{eq:wildnav_policy}
\end{equation}

\begin{figure*}[!t]
\centering
    \includegraphics[width=1.00\linewidth]{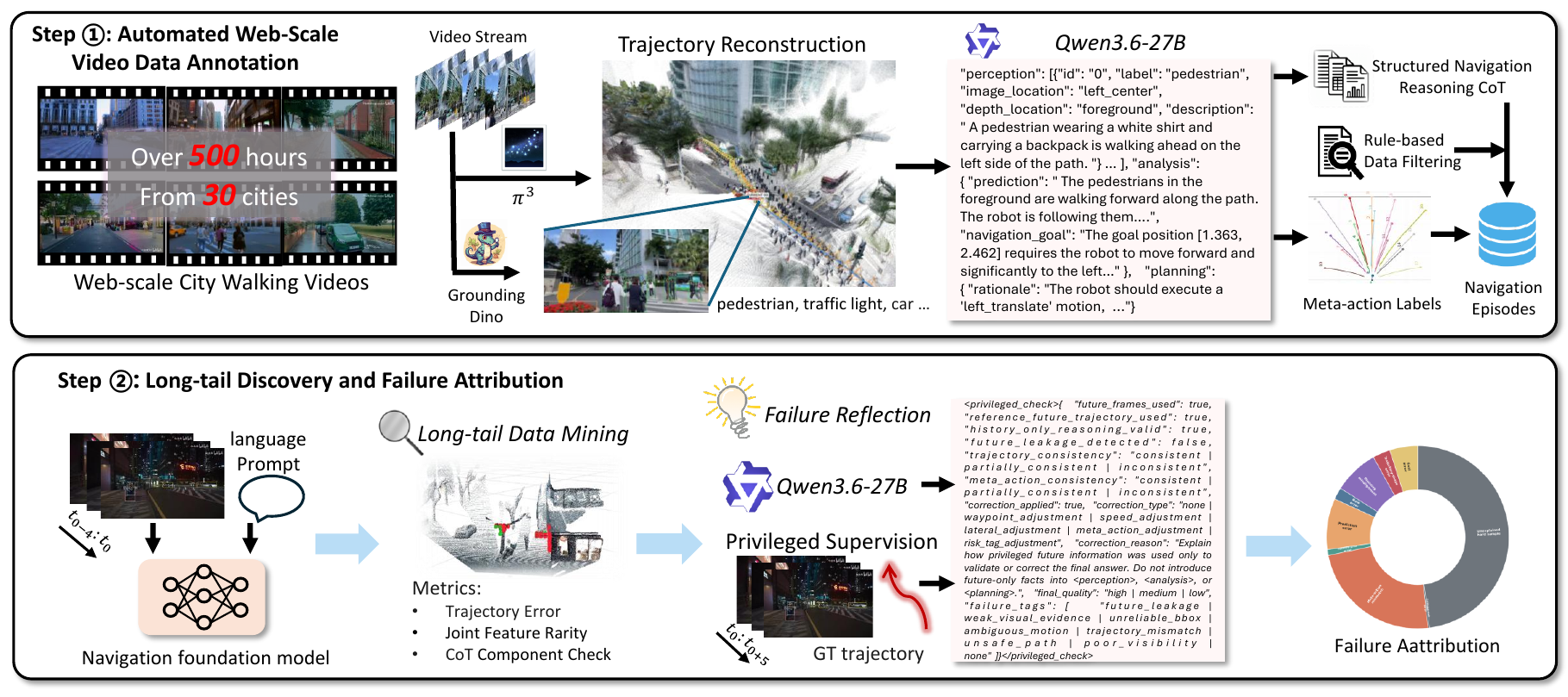}
    \caption{
The two-stage pipeline for automated video processing. Step 1 illustrates the annotation process consisting of Qwen3.6VL-27B and other visual foundation models. Step 2 depicts the long-tail data analysis process. Long-tail samples are defined using distributional rarity and model-dependent trajectory error, and a teacher model performs reflective reasoning for structured failure attribution.
    }
    \vspace{-0.5cm}
    \label{fig:data-annotation}
\end{figure*}

\subsection{From In-the-Wild Videos to Navigation Episodes}
\label{sec:annotation}

\paragraph{Web-scale Video Dataset.}
As shown in Fig.~\ref{fig:video_coverage}, our street-walking video collection spans 30 cities across 15 countries, providing broad coverage of urban appearance, pedestrian behavior, traffic and social conventions, weather, illumination, crowd density, and road layout.
We resize these in-the-wild egocentric videos to a uniform resolution and segment them into 5-minute clips. 
In-the-wild videos can contain clips that are outside the scope of our generic ground-plane navigation setting. 
We therefore use a vision-language model to identify video cuts, subject-focused clips, as well as special events such as doorway, stair, and elevator transitions.
These clips are filtered out of our urban navigation dataset.

\paragraph{Trajectory reconstruction.}
We reconstruct each standardized 5-minute clip using LoGeR, a long-context geometric reconstruction framework, with the Pi3X model as the visual geometry backbone \cite{zhang2026loger, wang2026pi3}. 
Pi3X estimates reference-free per-frame camera poses and metric-scale depth, while LoGeR aligns overlapping chunks with memory to preserve geometric consistency.
The reconstructed camera centers are aligned to a stable initial reference frame and projected onto the ground plane to form a continuous trajectory. 
To use the reconstructed walking trajectories as pedestrian demonstrations for the point-goal navigation task, each 5-minute clip is sampled at 1 Hz and segmented into 10-frame navigation episodes comprising 5 historical and 5 future frames.
Consecutive episodes are grouped into temporal blocks before dataset splitting so that strongly overlapping windows do not cross data partitions.

\paragraph{Meta-action Semantic Labels.}
High-level meta-actions provide a compact intermediate representation between semantic reasoning and metric waypoint planning, as adopted in recent navigation VLA models \cite{s2e, impromptu}. 
We map each valid future trajectory to one of 20 meta-actions using deterministic geometric rules based on its speed, lateral progress, and motion direction. 
After excluding the invalid episodes, the resulting label provides a categorical summary that remains consistent with the continuous waypoint trajectory.

% We derive each meta-action from the geometry of its ground-truth future trajectory using deterministic rules based on path length, final displacement, and lateral progress. 
% Trajectories containing pose discontinuities, implausible path lengths, or inconsistent motion are treated as invalid episodes and excluded from the dataset. 
% For each remaining trajectory, path length and final displacement specify the motion direction (forward or backward) and speed level (stop, slow, normal, or fast), while relative longitudinal and lateral progress classifies the path as straight, turning left or right, or predominantly lateral translation.
% These rules define 20 meta-action labels.
% The meta-action therefore serves as a discrete categorical summary of the continuous waypoint trajectory while remaining geometrically consistent with it.

\paragraph{Structured Navigation Reasoning Annotations.}

The reconstructed trajectories provide action supervision but do not explain why the demonstrated path is appropriate.
We first use GroundingDINO   \cite{liu2024grounding} to detect navigation-relevant objects and map the detection coordinates and corresponding reconstructed depth values to normalized object class, image-plane location, and depth labels.
Qwen3.6VL-27B \cite{bai2025qwen3vl} is employed as an annotation teacher VLM to generate structured navigation rationales.
It receives the historical and current observations and the point goal as the input observable evidence. 
The annotation follows a perception-analysis-planning schema \cite{autodrivep3}. 
The grounded perception labels are organized into the perception field.
The analysis field describes the predicted scene evolution, the goal location, feasible traversable space, applicable interaction rules, and potential collision risks. 
The planning field answers a rationale for the planned meta-action and trajectory. 
The teacher VLM additionally receives the future observations and rule-derived meta-actions, used only as privileged verification evidence.
The teacher is explicitly instructed to ground its rationale in cues observable from the historical and current inputs, preventing future-only events from appearing in the training annotation. 
Finally, each annotation is then validated for schema completeness, meta-action labels, and the ground-truth future trajectory.
% The perception field describes navigation-relevant instances with image-plane and depth locations. 
% For supervised learning, we retain the teacher's analysis fields and planning rationale as the textual target $c_t$. 
% The discrete meta-action and continuous waypoints are removed from the language target and supervised through dedicated planning heads. 
% The richer perception annotation remains attached to each episode and is later used to construct the semantic feature space for long-tail analysis. 
% This separation avoids asking the language decoder to encode metric coordinates as text while preserving a structured and auditable representation of scene semantics.

% Low-rank adapters with rank 32, scaling factor 64, and dropout 0.08 are inserted into the attention projections and feed-forward layers of the language model, allowing the backbone to acquire navigation-specific reasoning with a small number of trainable parameters.
\subsection{WILD-Nav Model}

WILD-Nav uses a Qwen3.5VL-4B model \cite{bai2025qwen3vl} as its backbone. 
The visual encoder converts the five sequential egocentric observations into image tokens, which are interleaved with a task prompt, alongside textual representations of the historical trajectory and point goal.
The visual encoder is frozen during training. 
To enable reasoning-aware trajectory planning, a special token, \texttt{<PLAN\_QUERY>}, is introduced as the interface for the action expert. 
Let $h_q\in\mathbb{R}^{2560}$ be the final normalized language-model hidden state corresponding to this token, which aggregates the preceding Chain-of-Thought (CoT) rationale. 
The action expert explicitly decodes $h_q$ into an executable continuous trajectory plan.
It consists of two lightweight output heads.
The meta-action head is a linear classifier:
\begin{equation}
\hat{\mathbf{a}}_t
=
\operatorname{softmax}
\left(
W_a h_q + b_a
\right),
\label{eq}
\end{equation}
where $\hat{\mathbf{a}}_t$ is the predicted class-probability distribution, and $W_a$ and $b_a$ are learnable parameters.
The trajectory head is a four-layer multilayer perceptron with three 512-dimensional hidden layers and GELU activations:
\begin{equation}
    \hat{\boldsymbol{\tau}}^{\mathrm{plan}}_t
    =
        f_{\mathrm{mlp}}(h_q).
\end{equation}

\begin{figure}[!t]
\centering
    \includegraphics[width=0.88\linewidth]{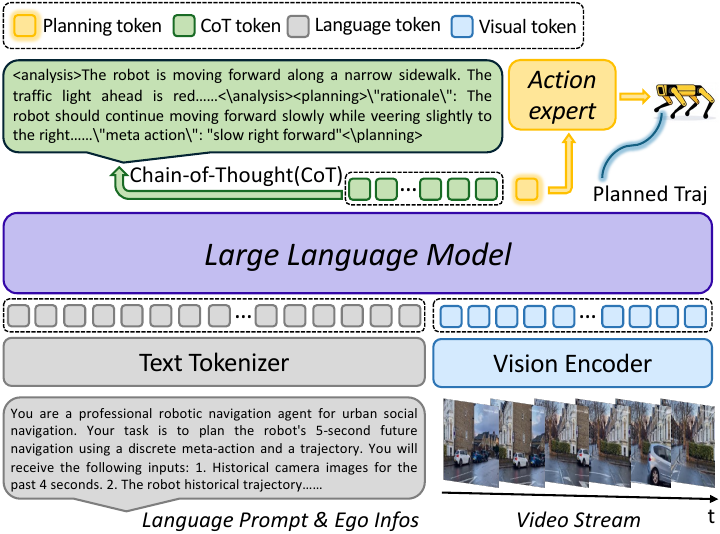}
    \caption{
    Detailed architecture of the proposed WILD-Nav model. It processes temporal video streams through a vision encoder, while a Large Language Model backbone processes the fused tokens to generate CoT reasoning. Finally, the action expert decodes the planning token into a trajectory.
    }
    \label{fig:vla-model}
    \vspace{-0.5cm}
\end{figure}

As illustrated in Fig.~\ref{fig:vla-model}, the pretrained vision-language backbone provides a shared reasoning representation, while the two lightweight heads decode it into the low-dimensional outputs required for precise navigation control.
In the primary reasoning-conditioned configuration, the model first autoregressively generates the structured rationale and terminates the assistant response.
The \texttt{\detokenize{<PLAN_QUERY>}} token is then appended after the completed rationale, and one additional forward pass produces the meta-action and waypoint plan.
Because $h_q$ attends to the complete generated rationale, the action expert is explicitly conditioned on the model's scene analysis.
For efficiency-oriented variants, \texttt{\detokenize{<PLAN_QUERY>}} can instead be placed before the rationale during training.
Causal attention prevents it from observing the subsequent text, allowing direct one-pass planning at inference while using the rationale as an auxiliary training objective.

\subsection{Multi-task Supervised Learning}

The model is optimized with three complementary objectives.
First, an autoregressive language-modeling loss supervises the structured analysis and planning rationale:
\begin{equation}
    \mathcal{L}_{\mathrm{cot}}
    =
    -\sum_{j\in\mathcal{I}_{\mathrm{ans}}}
    \log
    p_{\theta}
    \left(
        c_{t,j}
        \mid
        c_{t,<j},
        I_t,
        \mathcal{I}^{\mathrm{hist}}_t,
        \boldsymbol{\tau}^{\mathrm{hist}}_t,
        \mathbf{g}_t
    \right),
    \label{eq:cot_loss}
\end{equation}
where $\mathcal{I}_{\mathrm{ans}}$ denotes the index set of the generated CoT reasoning tokens.
The loss is computed only over the generated rationale, while the input prompt, padding positions, and the special \texttt{\detokenize{<PLAN_QUERY>}} token are masked.

Second, the meta-action head is trained using cross-entropy:
\begin{equation}
\mathcal{L}_{\mathrm{act}}
=
-\log \hat{\mathbf{a}}_t[a_t],
\label{eq}
\end{equation}
where $a_t$ is the ground-truth meta-action label and $\hat{\mathbf{a}}_t[a_t]$ is its predicted probability.

Third, the trajectory head is supervised using a Mean Squared Error (MSE) loss over the five future waypoints:
\begin{equation}
    \mathcal{L}_{\mathrm{traj\_mlp}}
    =
    \frac{1}{5}
    \sum_{k=1}^{5}
    \left\|
        \hat{\mathbf{p}}_{t+k} - \mathbf{p}_{t+k}
    \right\|_2^2.
    \label{eq:trajectory_loss_mlp}
\end{equation}
where $\hat{\mathbf{p}}_{t+k}$ and $\mathbf{p}_{t+k}$ denote the predicted and ground-truth waypoints, respectively.

The complete objective is
$
\mathcal{L}
=
\lambda_{\mathrm{cot}}\mathcal{L}_{\mathrm{cot}}
+
\lambda_{\mathrm{act}}\mathcal{L}_{\mathrm{act}}
+
\lambda_{\mathrm{traj}}\mathcal{L}_{\mathrm{traj}},
\label{eq}
$
where $\lambda_{\mathrm{cot}}$, $\lambda_{\mathrm{act}}$, and $\lambda_{\mathrm{traj}}$ control the relative weights of the three objectives.
The action expert uses a higher learning rate than the language-model adapters, while the visual encoder remains fixed.

\section{Long-Tail Discovery}
\label{sec:long_tail}

\subsection{Characterizing the Urban Navigation Long Tail}
The internet-sourced, web-scale videos serve both as navigation supervision and as an empirical basis for revealing its long tail beyond predefined scenario taxonomies. 
We represent each episode by a perception--motion pattern $z=\phi(x,y)$, where $x$ contains the visual observations, motion history, and point goal, while $y$ describes the meta-action and future trajectory. 
Let $p(z)$ denote its occurrence probability in the corpus and $r_f(z)=\mathbb{E}[\ell(f(x),y)\mid Z=z]$ the conditional planning risk of model $f$. 
The average risk is
\begin{equation}
    R(f)=\sum_z p(z)\,r_f(z).
    \label{eq:risk_decomposition}
\end{equation}
This decomposition explains that a low-frequency pattern can have high conditional error while contributing little to the average risk.
We therefore characterize the navigation long tail along two complementary axes: rarity measures the empirical support of a perception--motion pattern, while difficulty measures the planning error of a particular model. 
In addition, reflection over difficult episodes examines why the model fails. 
These statistics characterize the empirical distribution of our web-video corpus.

\subsection{Navigation-Conditioned Rarity}

Rarity is estimated from the joint perception and motion configuration.
Open-vocabulary object descriptions are mapped to a normalized vocabulary through lexical canonicalization.
The perception representation contains normalized object categories together with their coarse image-plane and depth locations.
The motion representation follows the same geometric rules used to construct the meta-action labels in Sec.~\ref{sec:web_video_learning}.

For episode $i$, we combine its object and spatial features with the corresponding motion configuration to obtain the joint feature. 
Feature rarity is measured using smoothed inverse document frequency and aggregated into an episode-level score:
\begin{equation}
\begin{aligned}
    \operatorname{IDF}(a)
    =
    \log\frac{N+1}{\operatorname{DF}(a)+1},
    s_i^{\mathrm{rare}}
    =
    \frac{\sum_{a\in\mathcal{F}_i}\operatorname{IDF}(a)}
    {\sqrt{|\mathcal{F}_i|}} .
\end{aligned}
\label{eq:rarity_score}
\end{equation}
$N$ is the total number of episodes, $\operatorname{DF}(a)$ is the number of episodes in which feature $a$ appears, $|\mathcal{F}_i|$ denotes the cardinality of all active joint features in the episode. 
The resulting score is higher when an episode contains more infrequent perception--motion combinations. 
This representation distinguishes common semantic elements from uncommon navigation configurations.

\subsection{Difficulty and Reflection-Based Attribution}

We estimate model-dependent difficulty using the Average Displacement Error (ADE) and Final Displacement Error (FDE) of planned trajectories. 
ADE measures the mean Euclidean displacement over all waypoints, while FDE measures the endpoint error relative to the local goal. 
We independently select the top-$3\%$ episodes under each metric and define their union as the hard set.

Privileged reflection is applied only to the hard set using the same teacher VLM as in Sec.~\ref{sec:annotation}. 
In addition to the historical observations and model outputs, the teacher model receives future observations and the reference trajectory for retrospective verification. 
It examines whether the scene analysis is supported by the observation history, whether the planning rationale agrees with the goal, and whether the trajectory is consistent with the meta-action. 
ADE, FDE, rarity scores, and hard-sample ranks are excluded from the reflection prompt so that the attribution remains independent of the mining criteria.

\section{Experiments}
\label{sec:experiments}

\subsection{Experimental Setup}
\label{sec:exp_setup}

\paragraph{Datasets.}
The collected in‑the‑wild video corpus is organized into two distinct subsets.
The first split provides the data for fine-tuning navigation models and is divided into 80\% training and 20\% test sets. 
The second part is reserved for long-tail discovery, including a mining set of 287820 samples and an evaluation set of 104655 samples. 
We further unify real-world navigation data from 3 datasets to evaluate cross-domain adaptation \cite{scand, sit, gnd}, which produces 45096 samples and is divided into train and test sets.
The original WILD-Nav models are trained only on the video training split, while real-world and long-tail data are introduced only in their corresponding fine-tuning experiments.

\paragraph{Baselines and model variants.}
We compare with UrbanNav~\cite{urbannav} and SocialNav~\cite{socialnav}, as they are the closest recent methods that learn visual urban navigation planning from scalable human navigation data. 
We denote the publicly released SocialNav weights as SocialNav-origin, and additionally fine-tune both UrbanNav and SocialNav on our training data.
The proposed method WILD-Nav is evaluated in two variants. 
WILD-Nav-inst directly computes the planning outputs from instruction-conditioned features, whereas WILD-Nav-think first generates the structured perception--analysis--planning rationale and then computes the planning outputs.
All experiments are conducted on NVIDIA V100 GPUs with 32 GB of memory.

\paragraph{Metrics.}
We report ADE and FDE to evaluate metric trajectory planning accuracy.
Following UrbanNav and SocialNav~\cite{urbannav,socialnav}, we additionally report maximum angular orientation error (MAOE), computed by averaging the largest waypoint-wise direction error of each episode. 
Meta-action accuracy is reported only for the WILD-Nav variants because the baselines do not support meta-action prediction.

\begin{table}[t!]
    \centering
    \caption{Comparison of navigation planning results on the video and real-world test sets. -- denotes the unavailable test.}
    \label{tab:navigation_results}
    \footnotesize
    \setlength{\tabcolsep}{2.3pt}
    \begin{tabular}{lrrrr}
        \toprule
        Method & Meta(\%) $\uparrow$ & ADE(m) $\downarrow$ & FDE(m) $\downarrow$ &
        MAOE($^\circ$) $\downarrow$ \\
        \midrule
        \multicolumn{5}{l}{\emph{Video Test Set}} \\
        UrbanNav & -- & 0.157 & 0.136  & 6.30 \\
        SocialNav-origin & -- & 0.906 & 1.123 & 16.53 \\
        SocialNav & -- & 0.470 & 0.728 & 12.65 \\
        WILD-Nav-inst & \textbf{96.5} & 0.094 & 0.071 & 5.90 \\
        WILD-Nav-think & 93.4 & \textbf{0.086} & \textbf{0.068} &
        \textbf{5.32} \\
        \midrule
        \multicolumn{5}{l}{\emph{Real-World Test Set: weights from video dataset}} \\
        UrbanNav & -- & 0.867 & 1.493 & 13.92 \\
        SocialNav & -- & 3.499 & 5.824 & 19.49 \\
        WILD-Nav-inst & \textbf{93.4} & 0.953 & 1.493 & 7.36 \\
        WILD-Nav-think & 93.3 & \textbf{0.619} & \textbf{0.834} &
        \textbf{7.14} \\
        \midrule
        \multicolumn{5}{l}{\emph{Real-World Test Set: fine-tuned by real-data}} \\
        UrbanNav & -- & 0.257  &  0.286 & 8.29  \\
        SocialNav & -- & 2.901 & 4.837 & 16.08 \\
        WILD-Nav-inst & 89.13 & \textbf{0.206} & \textbf{0.127} &
        \textbf{4.49} \\
        \bottomrule
    \end{tabular}
\vspace{-0.2cm}
\end{table}

\subsection{Navigation Planning Results}
\label{sec:navigation_results}

\paragraph{Video Test Set.}
We first evaluate urban navigation VLA models using in-the-wild egocentric videos to examine whether such videos can serve as scalable human navigation demonstrations.
WILD-Nav consistently outperforms both baselines across all trajectory metrics. 
Even compared with the stronger SocialNav variant, WILD-Nav-inst reduces all three planning errors by at least 53.4\%, while achieving a 47.8\% reduction in FDE over UrbanNav.
WILD-Nav-think further improves upon WILD-Nav-inst, reducing ADE and FDE by 8.5\% and 4.2\%, respectively. 
The comparisons reveal the respective contributions of CoT data and model design.
The further advantage of WILD-Nav over the baselines mainly shows its improved model design.
WILD-Nav-think performs best across all trajectory metrics, suggesting that structured reasoning provides useful context for continuous waypoint generation.

\paragraph{Real-World Test Set.}
We evaluate two settings on the Real-World Test Set: direct transfer using weights learned from the video dataset, and domain adaptation by fine-tuning the models on the real-world training split. 
The directly transferred models exhibit substantially higher errors, indicating a considerable domain gap caused by differences in camera viewpoints and motion constraints. 
After real-world fine-tuning, WILD-Nav-inst achieves the best trajectory performance, reducing FDE and MAOE by 55.6\% and 45.8\%, respectively, compared with the next-best UrbanNav. 
Nevertheless, its ADE and FDE remain higher than those on the Video Test Set, suggesting that fine-tuning does not completely close the domain gap. 
A likely reason is that the real-world training set is substantially smaller than the video dataset and therefore covers fewer visual and motion patterns.

\begin{figure}[t!]
    \centering
    \begin{subfigure}[b]{0.23\textwidth}
        \centering
        \includegraphics[width=\textwidth]{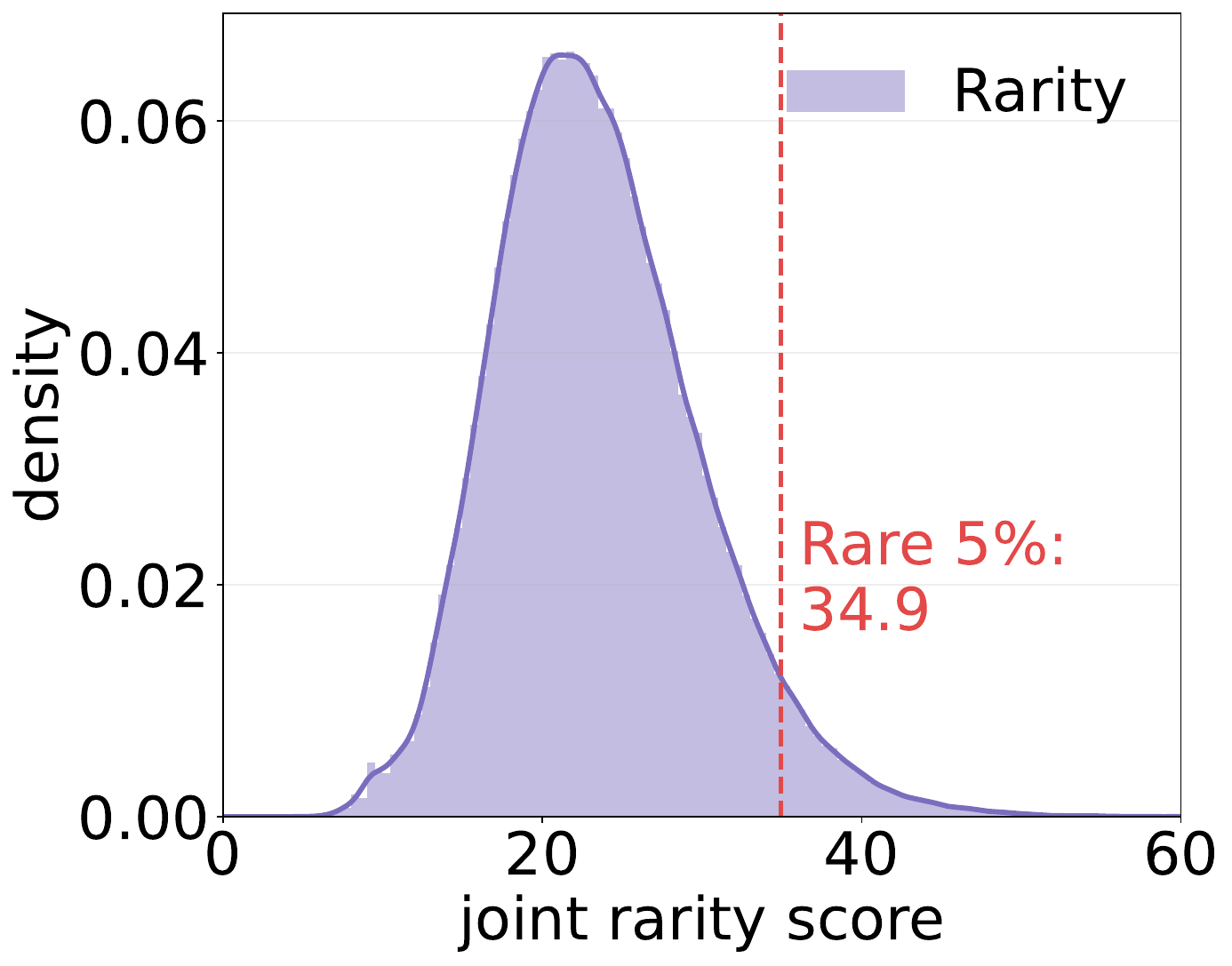}
    \end{subfigure}
    \begin{subfigure}[b]{0.23\textwidth}
        \centering
        \includegraphics[width=\textwidth]{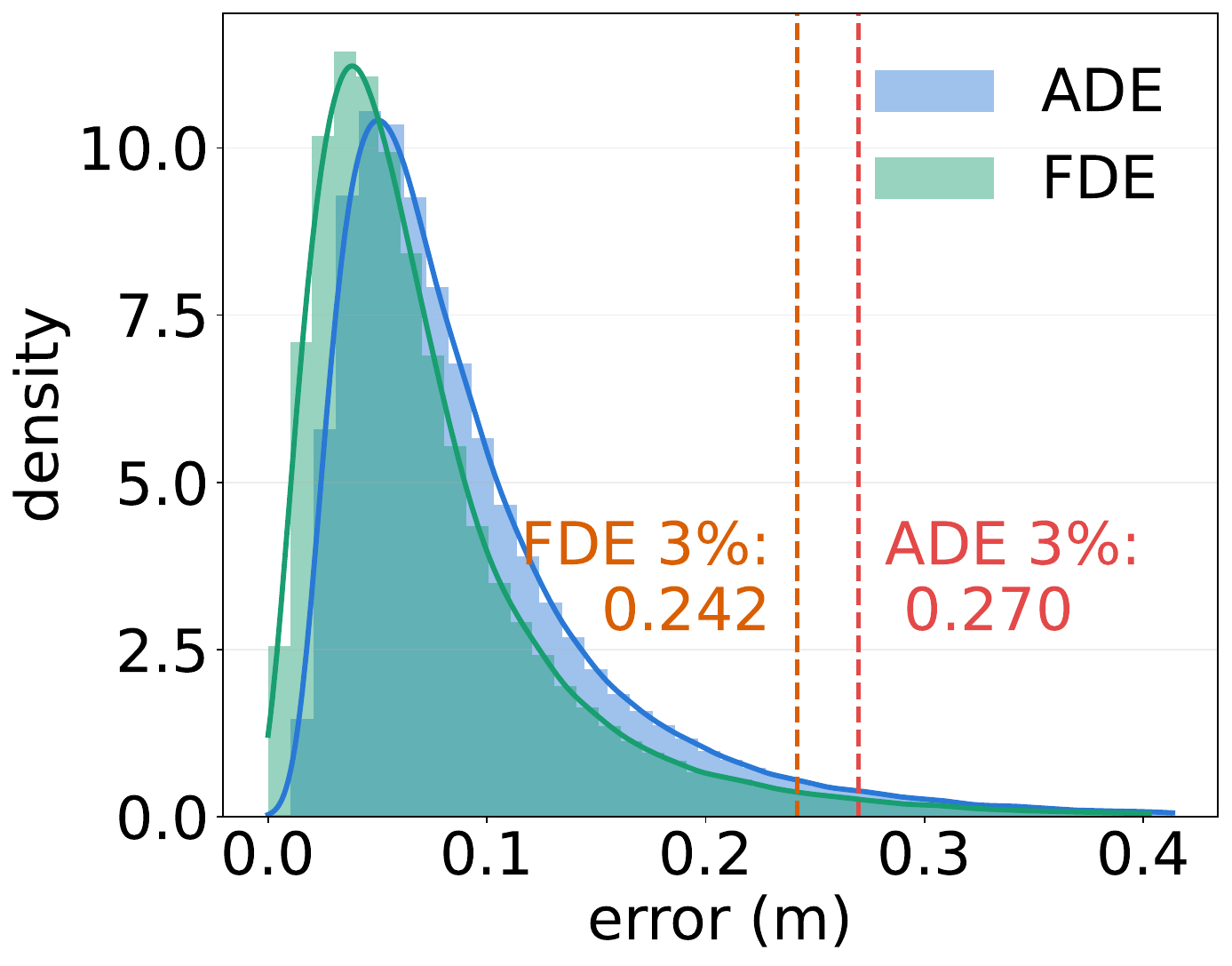}
    \end{subfigure}
    % \begin{subfigure}[b]{0.17\textwidth}
    %     \centering
    %     \includegraphics[width=\textwidth]{Figures/hard_rare_coverage.pdf}
    %     \caption{}
    % \end{subfigure}
    \caption{
    Overview of long-tail discovery through feature rarity and model-dependent difficulty.
    Left: Distribution of joint perception--motion rarity scores.
    Right: Distributions of ADE and FDE.
    % (c) Coverage between hard and rare episodes.
    % (d) Primary failure attribution obtained by privileged reflection over the hard set.
    }
    \label{fig:longtail}
\vspace{-0.5cm}
\end{figure}

\subsection{Long-Tail Evaluation Protocol}
\label{sec:long_tail_experiments}

Figure~\ref{fig:longtail} summarizes the rarity and model-error distributions, their overlap, and reflection-based failure attribution. 
Both the rarity score and error distributions show a right-tailed distribution as depicted in Figure~\ref{fig:longtail}.
We operationally define the samples with top-$5\%$ rarity scores as rare perception--motion patterns. 
Difficulty is defined according to the performance of WILD-Nav-inst, yielding a hard set that covers 4.92\% of the whole Long-Tail Mining Set.
% Figure~\ref{fig:longtail} (c) shows that rarity and difficulty are associated but distinct.
For the coverage between them, 24.18\% of hard episodes are rare, while 14.48\% of rare episodes are hard. 
Their joint analysis therefore distinguishes common--hard cases revealing systematic model limitations, and rare--hard cases combining weak data support with high model error.

\begin{figure}[t!]
    \centering
    \begin{subfigure}[b]{0.2\textwidth}
        \centering
        \includegraphics[width=\textwidth]{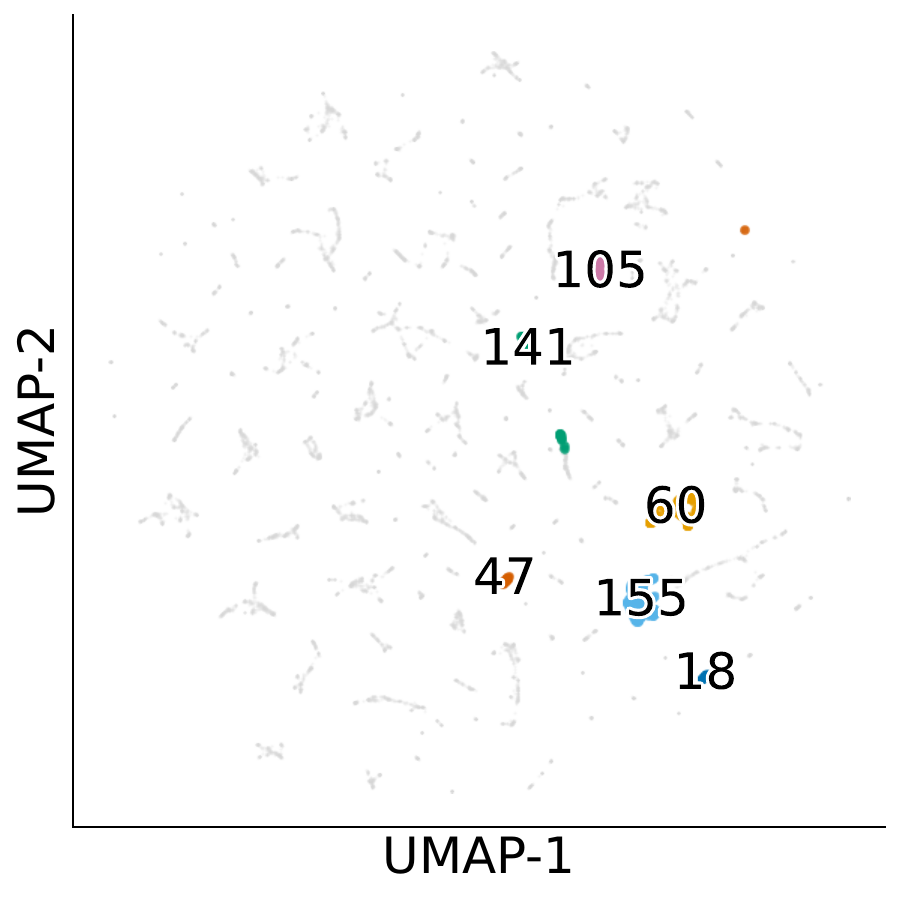}
        \caption{}
    \end{subfigure}
    \begin{subfigure}[b]{0.22\textwidth}
        \centering
        \includegraphics[width=\textwidth]{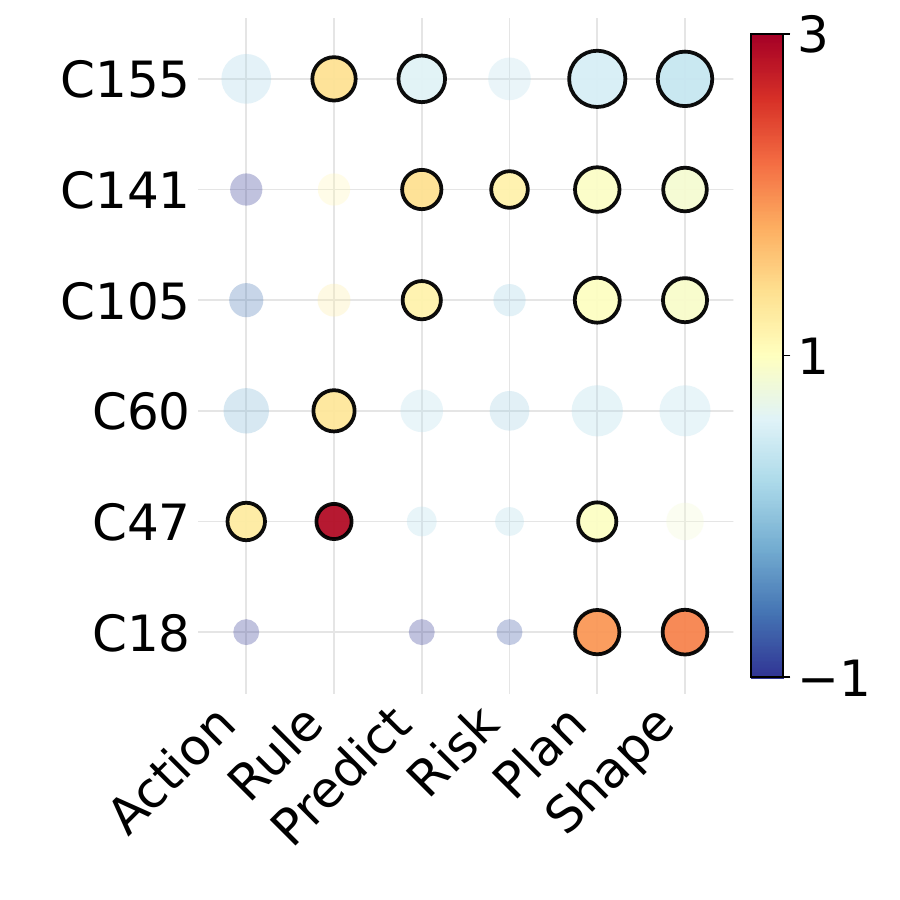}
        \caption{}
    \end{subfigure}
    
    % \hfill
    % \begin{subfigure}[b]{0.22\textwidth}
    %     \centering
    %     \includegraphics[width=\textwidth]{Figures/primary_reflection_failure_donut.pdf}
    %     \caption{}
    % \end{subfigure}
    \caption{
    Failure attribution obtained by privileged reflection over the hard set. 
    (a) Visualization of six representative HDBSCAN clusters in the hard set; gray points denote all remaining samples.
    (b) Cluster-specific reflection failure patterns. Bubble size represents failure support, and color indicates enrichment over the global baseline.
    }
    \label{fig:fail}
\vspace{-0.3cm}
\end{figure}

% Reflection provides a complementary diagnosis of these failures. 
% While 56.87\% of hard episodes have no clear reasoning-stage attribution, meta-action mismatch is the largest explicit cause, accounting for 42.63\% of attributed episodes. 
% Multi-label results further show that planning misalignment and trajectory-shape errors frequently co-occur, suggesting that navigation failures often propagate across multiple decision stages.
To diagnose the failure mechanisms of difficult samples, we first cluster the samples using the joint feature and associate each cluster with the reflection failures. 
The joint features were first compressed using SVD and mapped to a neighborhood-preserving latent space using UMAP. 
HDBSCAN was then applied to discover variable-density scene clusters.
Fig. \ref{fig:fail} (a) visualizes six representative scene clusters, while (b) reports their failure enrichment relative to the global baseline. 
Distinct failure patterns emerge: dense-pedestrian stopping scenes (C18) are dominated by planning and trajectory-shape errors; scooter–sign contexts (C47) exhibit rule and action errors; traffic-sign scenes (C60) amplify rule violations; and scooter or cyclist interactions (C105/C141) primarily expose prediction and collision-risk failures. 
These results indicate that model failures are scene-dependent rather than uniformly distributed. 
The identified patterns should be interpreted as statistically supported local associations rather than exhaustive causal categories.

\begin{table}[t!]
    \centering
    \caption{Ablation of incremental fine-tuning data on the Long-Tail Test Set. The fixed tail is the union of the worst 3\% ADE and FDE samples selected by WILD-Nav-inst. }
    \label{tab:hard_finetuning}
    \footnotesize
    \setlength{\tabcolsep}{1.5pt}
    \begin{tabular}{lrrrrr}
        \toprule
        & \multicolumn{2}{c}{Test-all} &
        \multicolumn{3}{c}{Test-hard} \\
        \cmidrule(lr){2-3}\cmidrule(lr){4-6}
        Method & ADE $\downarrow$ & FDE $\downarrow$ &
        ADE $\downarrow$ & FDE $\downarrow$ & Meta (\%) $\uparrow$ \\
        \midrule
        WILD-Nav-inst & 0.094 & 0.074 & 0.286 & 0.247 & 94.88 \\
        \quad +Random & \textbf{0.091} & \textbf{0.072} &
        0.277 & 0.231 & 95.10 \\
        \quad +Hard & 0.121 & 0.102 & 0.230 & 0.165 &
        \textbf{96.83} \\
        \quad +Hard+Random & 0.093 & 0.079 & \textbf{0.217} &
        \textbf{0.160} & 96.73 \\
        \bottomrule
    \end{tabular}
\vspace{-0.5cm}
\end{table}

\subsection{Results on the Long-Tail Test Set}
\label{sec:longtail_results}

We evaluate whether model-dependent hard examples provide more effective fine-tuning supervision than randomly sampled data. 
For evaluation, the same rule is applied once to the predictions of WILD-Nav-inst on the Long-Tail Test Set. 
All fine-tuned models are evaluated on these same sample IDs without redefining the tail.
Table~\ref{tab:hard_finetuning} compares the base model with three incremental fine-tuning strategies. 
Random fine-tuning mainly preserves overall performance but provides limited gains on the fixed tail. 
Hard fine-tuning improves tail performance more effectively but degrades the full test set. 
Further adding random samples provides the best balance by preserving overall performance while achieving the lowest tail errors. 
This indicates that hard examples provide targeted supervision for failure cases, whereas replay helps retain the broader navigation distribution.

% TODO: Insert the rarity--difficulty quadrant statistics and the
% reflection-based failure-stage distribution once their audited aggregate
% results are available.

\begin{figure}[t!]
\centering
    \includegraphics[width=0.95\linewidth]{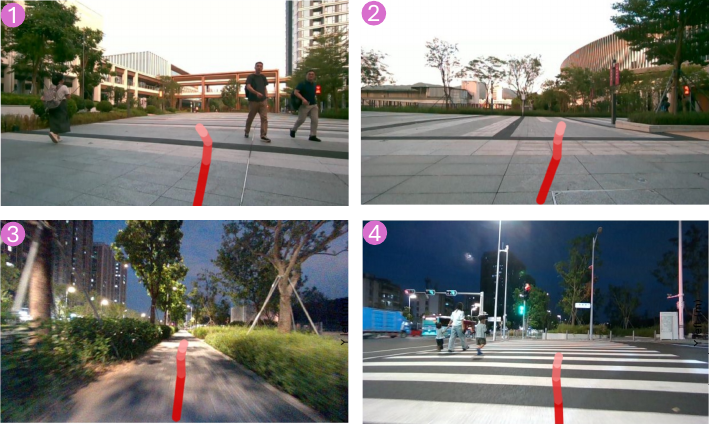}
    \caption{
    Real-world navigation performance of our VLA model. Our model generates safe and reasonable trajectories (red) for various navigation tasks.
    }
    \label{fig:realworld}
\vspace{-0.4cm}
\end{figure}
% including (1) yielding to dynamic pedestrians, (2) cornering at intersections, (3) avoiding obstacles on sidewalks, and (4) obeying traffic signals at crosswalks

\paragraph{Qualitative Real-World Results.}
We deploy the fine-tuned WILD-Nav-inst model on a mobile robot and conduct navigation trials in diverse outdoor urban environments.
Figure~\ref{fig:realworld} presents representative planning results. 
Across pedestrian interactions, intersection turns, sidewalk obstacles, and signalized crossings, WILD-Nav generates trajectories that remain within traversable regions or surrounding constraints. 
More details are in the Appendix.

\section{Conclusion}
We presented a scalable framework that converts 500 hours of in-the-wild street-walking videos across 30 cities into more than 500K navigation samples, and used these data to train WILD-Nav, a reasoning-aware VLA model for point-goal urban navigation.
It achieves accurate and interpretable point-goal planning on both video and real-world data. 
Beyond aggregate evaluation, our joint analysis of perception--motion rarity and model-dependent difficulty exposes distinct long-tail structures, while privileged reflection attributes difficult cases to recurring reasoning and planning failures. 
Targeted fine-tuning further improves performance on the fixed tail, with replay data helping preserve overall capability. 
These results establish in-the-wild videos as both scalable navigation supervision and an empirical basis for identifying the data and model bottlenecks that limit urban navigation.

\bigskip
% \noindent Thank you for reading these instructions carefully. We look forward to receiving your electronic files!

\bibliography{aaai2027}


\end{document}